\documentclass[review]{elsarticle}
\usepackage{caption}
\usepackage{color}
\usepackage{multirow}
\usepackage{hyperref}

\usepackage{amsmath}
\usepackage{booktabs}
\usepackage[subrefformat=parens,labelformat=parens]{subfig}
\usepackage{amssymb}
\usepackage{arydshln}  
\usepackage{float}  
\usepackage[all]{nowidow}
\usepackage{longtable}
\usepackage[title]{appendix}
\usepackage{bm}
\usepackage[table]{xcolor}
\usepackage{hyperref}

\usepackage[whole]{bxcjkjatype}

\journal{Pattern Recognition}

\biboptions{numbers,sort&compress}
\begin{document}

\begin{frontmatter}
\title{Learning from Majority Label: A Novel Problem in Multi-class Multiple-Instance Learning}


\author[kyushu_univ]{Shiku~Kaito\corref{mycorrespondingauthor}}
\cortext[mycorrespondingauthor]{Corresponding author}
\ead{kaito.shiku@human.ait.kyushu-u.ac.jp}

\author[kyushu_univ]{Shinnosuke~Matsuo}
\author[kyushu_univ]{Daiki~Suehiro}
\author[kyushu_univ]{Ryoma~Bise}
\ead{bise@ait.kyushu-u.ac.jp}

\address[kyushu_univ]{Department of Advanced Information Technology, Kyushu University, Fukuoka, Japan}

\begin{abstract}
The paper proposes a novel multi-class Multiple-Instance Learning (MIL) problem called Learning from Majority Label (LML). 
In LML, the majority class of instances in a bag is assigned as the bag-level label. 
The goal of LML is to train a classification model that estimates the class of each instance using the majority label.
This problem is valuable in a variety of applications, including pathology image segmentation, political voting prediction, customer sentiment analysis, and environmental monitoring.
To solve LML, we propose a Counting Network trained to produce bag-level majority labels, estimated by counting the number of instances in each class.
Furthermore, analysis experiments on the characteristics of LML revealed that bags with a high proportion of the majority class facilitate learning. Based on this result, we developed a Majority Proportion Enhancement Module (MPEM) that increases the proportion of the majority class by removing minority class instances within the bags.
Experiments demonstrate the superiority of the proposed method on four datasets compared to conventional MIL methods. Moreover, ablation studies confirmed the effectiveness of each module.
The code is available at \href{https://github.com/Shiku-Kaito/Learning-from-Majority-Label-A-Novel-Problem-in-Multi-class-Multiple-Instance-Learning}{here}.
\end{abstract}

\end{frontmatter}

\section{Introduction\label{sec:intro}}

Multiple-Instance Learning (MIL)~\cite{ramon2000multi, wang2018revisiting,Ilse2020DeepMI, Pinheiro2015,ilse2018attention, shao2021transmil,rymarczyk2021kernel,zhou2012multi,yang2017miml,feng2017deep} is a widely studied machine learning task, with various applications, including video recognition~\cite{ren2023proposal, arnab2021vivit, lv2023unbiased, chen2024prompt}, and pathology image classification~\cite{ilse2018attention, shao2021transmil, rymarczyk2021kernel, javed2022additive, qu2023boosting, tang2023multiple, chen2023rankmix}.
In traditional supervised learning, each data point (instance) is assigned a label, and the goal is to train a model to classify each instance. On the other hand, in MIL, individual instances do not have labels. Instead, a group of instances, called a `bag,' is assigned a label determined by certain rules or patterns related to the instances within the bag, and the goal is to train a classifier at the bag level.

In MIL, the relationship between a bag and its instances is often determined by specific, context-dependent rules that determine how the bag's label is assigned based on the instance labels. These rules can vary widely depending on the nature of the task and the data.
In binary MIL~\cite{shao2021transmil, javed2022additive, rymarczyk2021kernel,early2024inherently}, a positive label is typically assigned to the bag if at least one instance within it is positive.
In multi-class MIL, the label assignment can be more complex. For example, in severity classification tasks, the instance with the highest severity level typically determines the bag’s overall label, reflecting the highest severity within the bag~\cite{Shiku_2025_WACV, schwab2022automatic}.
Despite the variety of these underlying rules, many existing MIL methods do not explicitly consider them. Instead, they often focus on aggregating instance-level features, using simple techniques like averaging~\cite{wang2018revisiting, Ilse2020DeepMI} or weighted sum~\cite{ilse2018attention}, sometimes with attention mechanism~\cite{javed2022additive, shao2021transmil}, without addressing how these rules influence the final bag label.


\begin{figure}[t]
\centering
\includegraphics[width=0.88\columnwidth]{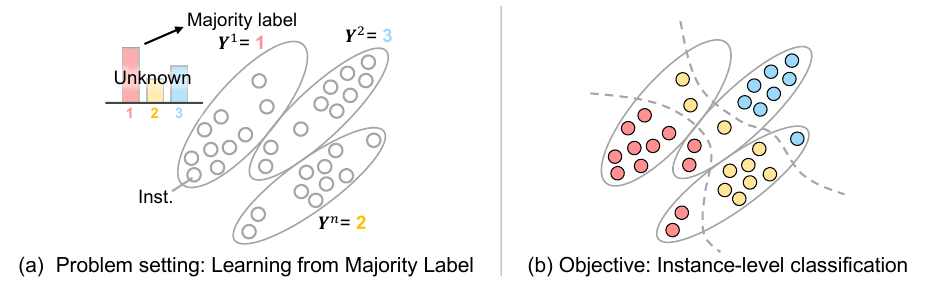}
\vspace{-3.5mm}
\caption{A novel problem setting: Learning from Majority Label (LML). The goal of LML is to train a classification model that estimates the class of each instance using the majority label. 
In this setting, while bag-level supervision is provided during training, instance-level supervision is not.
}
\label{fig:intro}
\vspace{-4.5mm}
\end{figure}

In this paper, we focus on one important relationship between the bag and instances in multi-class classification: Learning from Majority Label (LML). In LML, the bag-level label is determined by the majority class of instances within the bag. 
As shown in Figure~\ref{fig:intro}, the goal of LML is to train a classification model that estimates the class of each instance using the majority label.
This approach is useful in scenarios where individual instance labels are difficult to obtain. For instance, in pathology image segmentation~\cite{ren2019pancreatic, travis2011international}, it facilitates the identification of the dominant cancer subtype. Moreover, in political voting prediction~\cite{mciver2017privacy}, it enables the estimation of voting preferences based on the overall winner. 
Further applications include customer sentiment analysis~\cite{no_label_no_cry} and environmental monitoring~\cite{wang2011mixture}. Although LML has not been widely explored in prior research, it provides a practical solution to multi-class MIL problems.

\begin{figure}[t]
\centering
\includegraphics[width=1\columnwidth]{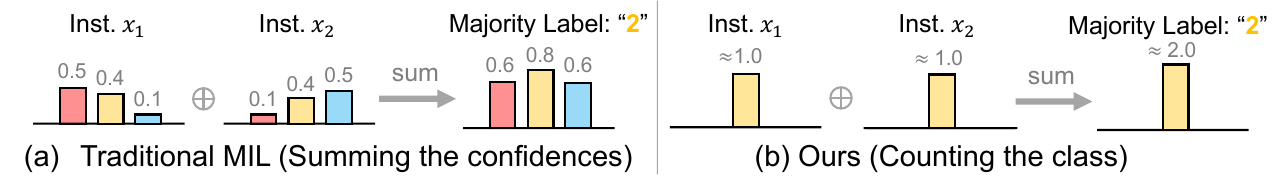}
\vspace{-8mm}
 \caption{Counting approach. (a) Inconsistent bag label determination by conventional MIL methods. (b) Consistent bag label determination by counting.}
\label{fig:counting}
\vspace{-3mm}
\end{figure}

In traditional MIL methods, determining the majority class can be ambiguous when instance-level predictions are aggregated, such as by summing class confidences~\cite{ wang2018revisiting}. For example, consider a bag with two instances and a majority class of 2. If the predictions are (0.5, 0.4, 0.1) and (0.1, 0.4, 0.5), the summed confidence is (0.6, 0.8, 0.6), making the majority class unclear. Similarly, predictions like (0.1, 0.4, 0.5) and (0.4, 0.6, 0) would result in (0.5, 1.0, 0.5), still ambiguous, as shown in Figure~\ref{fig:counting}(a). These multiple possible solutions complicate the training process and hinder accurate majority class identification.

To address these issues, we propose a Counting Network that estimates the majority class by directly counting the predicted classes of instances within a bag. The network forces class confidences to be binary-like, with values close to 1 for the predicted class and close to 0 for others. This constraint narrows the solution space, stabilizing training by avoiding bad local minima. In contrast to traditional methods, our Counting Network removes ambiguity by directly counting the predicted classes. For example, if the network predicts (0, 1, 0) for both instances, the summed prediction is (0, 2, 0), clearly identifying class 2 as the majority, as shown in Figure~\ref{fig:counting}(b). This approach ensures consistent majority class identification and avoids the inconsistencies found in traditional MIL methods.

\begin{figure}[t]
\centering
\includegraphics[width=0.5\columnwidth]{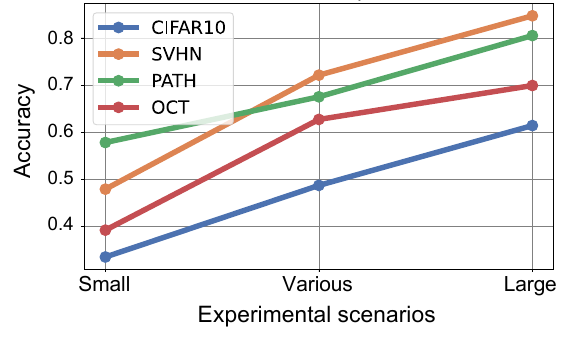}
\vspace{-2mm}
\caption{
Instance-level accuracy when changing the proportion of majority-class instances in a bag; {\bf Small} (greater than $1/C$ and up to $0.4$), {\bf Various} (greater than $1/C$ to $1$), and {\bf Large} ($0.6$ to $1$), where $C$ denotes the number of classes.
This result indicates that model performance improves when the proportion of majority-class instances in a bag is high.
}
\label{fig:Preliminary_analysis}
\vspace{-3mm}
\end{figure}

Additionally, we investigate the impact of the majority class proportion in the bag on learning performance of 
the Counting Network. 
Figure~\ref{fig:Preliminary_analysis} shows the instance-level classification accuracy when changing the proportion of majority-class instances in a bag; {\bf Small} (greater than $1/C$ and up to $0.4$), {\bf Various} (greater than $1/C$ to $1$), and {\bf Large} ($0.6$ to $1$), where $C$ denotes the number of classes.
Specifically, we observed that when the proportion of the majority-class instances in a bag is high, the model tends to perform better at instance-level classification. This insight is crucial, as the variance in the majority class proportion within bags can significantly affect how the classifier learns to generalize.


Based on this observation, we introduce the Majority Proportion Enhancement Module (MPEM)\footnote{The basic idea of the Counting Network was proposed in ICASSP~\cite{shikuIcassp}, and in this journal version, we extend it by introducing MPEM and conducting a more extensive experimental evaluation.}.
The purpose of MPEM is to increase the proportion of the majority class within a bag, thereby reducing the risk of misclassifying instances from the majority class as belonging to the minority class (i.e., non-majority class).
Specifically, when the proportion of the majority class in a bag is low, there is a higher risk that instances from the minority class may be incorrectly classified as belonging to the majority class.
By increasing the majority class proportion, MPEM mitigates this risk, improving the model's ability to generalize by ensuring that the classifier learns with a more balanced representation of the majority class.

In experiments conducted on four datasets, our proposed method, including the MPEM module, outperformed conventional MIL approaches. Additionally, an ablation study showed that the Counting Network improves consistency between the majority class predicted by the network and the one obtained by counting instances. 
It was confirmed that MPEM increases the majority class proportion in the bag and improves instance-level classification.
\section{Related Work\label{sec:related}}

\subsection{Bag-Level Classification in Multiple-Instance Learning (MIL)}
Conventional MIL primarily focuses on bag-level classification, with various methods proposed to aggregate instance-level predictions for bag-level classification. These methods can be broadly categorized into two types: output aggregation~\cite{ramon2000multi, wang2018revisiting,javed2022additive,early2024inherently} and feature aggregation~\cite{Ilse2020DeepMI, Pinheiro2015,ilse2018attention, shao2021transmil, rymarczyk2021kernel}.

In output aggregation, class confidences are predicted for each instance using an instance-level classifier, which are then aggregated (e.g., summed) to obtain a bag-level prediction. The loss, such as cross-entropy, is then computed between the predicted bag-level class confidence and the true bag-level label. In feature aggregation, instance-level features, obtained from an instance-level feature extractor, are aggregated into a bag-level feature. This bag-level feature is subsequently passed through a bag-level classifier to obtain the bag-level prediction.

Various aggregation methods, such as mean~\cite{wang2018revisiting, Ilse2020DeepMI}, max~\cite{Ilse2020DeepMI}, P-norm, and log-sum-exponential (LSE)~\cite{Pinheiro2015}, have been proposed to improve these aggregation methods. State-of-the-art approaches also leverage attention mechanisms~\cite{shao2021transmil, javed2022additive, rymarczyk2021kernel, early2024inherently} to weigh the importance of individual instances during aggregation.

Other approaches have introduced techniques for enhancing training, such as bag-level data augmentation~\cite{liu2024pseudo, gadermayr2023mixup, chen2023rankmix, yang2022remix, li2021novel}, hard instance mining~\cite{li2019deep, tang2023multiple}, and positive instance ratio boosting~\cite{qu2023boosting}.

However, LML differs from conventional MIL in that it focuses on instance-level classification and requires accurate instance counting, which is difficult with weighted aggregation methods. Additionally, in LML, the bag-level label is determined by the majority class of instances, which complicates the direct use of traditional MIL aggregation methods.

\subsection{Instance-Level Classification in MIL\label{subsec:dtw_and_ml}}
Instance-level classification has been less explored in MIL, which traditionally focuses on bag-level classification. Some recent studies~\cite{qu2022dgmil, as2024sm, qu2022bi} have investigated instance-level classification in the conventional binary MIL setting, such as cancer vs. non-cancer region classification in Whole Slide Image (WSI) segmentation. For example, DGMIL~\cite{qu2022dgmil} assumes that negative bags contain only negative instances, assigning pseudo-labels to instances in positive bags for instance-level classification. Similarly, attention-based methods~\cite{as2024sm, qu2022bi} have been used to identify important instances within a bag for instance-level classification.

However, these methods focus on the binary MIL setting, where the classification task is simpler and primarily aims to identify a small number of important instances within the bag. These approaches are not suitable for LML, which deals with multi-class classification, where the goal is to identify the majority class of instances within a bag.

To address instance-level classification in MIL with bag-level labeled data, multi-label MIL~\cite{zhou2012multi, yang2017miml, feng2017deep} methods have been employed. In multi-label MIL, each bag is assigned a set of labels, where each label indicates whether a specific class is present or absent within the bag. This allows the model to handle situations where a bag can contain instances of multiple classes simultaneously.
However, multi-label MIL is not suitable for LML because it assigns multiple labels to each bag, reflecting the presence or absence of several classes. In contrast, LML assigns a single label based on the majority class in the bag, making it unknown whether other classes are present or not. This fundamental difference in labeling makes multi-label MIL incompatible with the majority-class focused framework used in LML.

\begin{table}[t]
\caption{Comparison between LML and conventional MILs in a multi-class setting. 
``Number of bag-level labels'' denotes the number of labels assigned to a bag. For conventional multi-label MIL, each class is associated with a separate binary label, e.g., ``Class A: present or not.''
``Bag-level label decision rule'' denotes the relationship between the bag-level label and the instance-level labels. Although the conventional single-label multi-class MIL setting has not been sufficiently explored, LML specifically focuses on the case where the bag-level label is determined by the majority class.
``Prior work'' refers to related studies in each setting.
\label{tab:related}}
\centering
\scalebox{0.7}[0.7]{
\begin{tabular}{l|c|c|c}
\hline
  & LML (Ours)    & \multicolumn{2}{c}{Conventional MIL}  \\ \hline \hline
 Number of bag-level labels & \multicolumn{2}{c|}{Single} & Multiple \\ \hline
 \multirow{3}{*}{Bag-level label decision rule} & \multirow{3}{*}{\textbf{Majority}} & \multirow{3}{*}{Not explored} &  Class A: present or not  \\
 &&&  Class B: present or not  \\ 
 &&&  Class C: present or not  \\ \hline
Prior work     & \textbf{None} & ~\cite{early2024inherently} etc. & \cite{zhou2012multi, yang2017miml, feng2017deep} etc.\\ \hline
\end{tabular}
}
\end{table}

\subsection{Summary of Settings in Multi-Class Instance-Level MIL\label{subsec:relationjship_of_multiclassMIL}}
Table~\ref{tab:related} summarizes the relationships among problem settings in multi‑class instance‑level MIL. Most existing MIL methods assume that a bag-level label simply indicates whether the bag contains at least one instance of a particular class~\cite{zhou2012multi, yang2017miml, feng2017deep}. Therefore, they assign a separate binary label to every class (Multi labels): if the label for class A is positive, nothing is said about whether the bag also contains an instance of class B.

In a single‑label, multi‑class MIL setting, various kinds of relationships can be defined between the bag-level label and the instance-level labels; for example, the majority class, the highest ordinal class~\cite{Shiku_2025_WACV, schwab2022automatic}, and other possibilities. However, this single‑label multi‑class formulation has been little explored, and the few existing studies do not explicitly model this rule~\cite{early2024inherently}.
Our research began with the question of whether focusing on a specific bag‑instance relationship could lead to more effective methods. We therefore concentrate on learning from majority labels, a practical scenario that arises widely in real‑world applications and thus has broad applicability.

\par
\section{Learning from Majority\label{sec:method}}
\subsection{Problem Setting of Learning from Majority}

Given a bag  consisting of instances 
$\mathcal{B}^i = \{\bm{x}_j^i\}_{j=1}^{|\mathcal{B}^i|}$ and the bag-level label $\bm{Y}^i$, which is the majority class label, our goal is to estimate the instance-level labels $\bm{y}_j^i \in \{0,1\}^C$, $j=1 ,..., |\mathcal{B}^i|$.
Here, $|\mathcal{B}^i|$ is the number of instances contained within a bag $\mathcal{B}^i$.
$\bm{Y}^i$ is a $C$-dimensional one-hot vector where the element representing the majority class is set to 1, and $C$ denotes the number of classes.
Note that instance-level labels are not given during training.


We organize the relationship between bag-level majority label $\bm{Y}^i$ and instance-level label $\bm{y}_j^i$.
The number of instances belonging to class $k$ within a bag $\mathcal{B}^i$ is defined as follows: 
\begin{equation}
\label{eq:countVec}
N^i_k=\sum\limits_{j=1}^{|\mathcal{B}^i|}{y_{j,k}^i}, \\
\end{equation}
where $y_{j,k}^i$ represents the $k$-th element (class $k$) of $\bm{y}_{j}^i$. 
The summation denotes the operation of counting instances. The counting vector $\bm{N}^i$ is defined as $\bm{N}^i=(N_1^i ,\ldots, N_k^i ,\ldots, N_C^i)^T$, which provides the count of instances for each class.

The majority class is obtained through the $\arg\max$ operation, which retrieves the index of the element with the maximum value in the counting vector 
$\bm{N}^i$.
The majority label $\bm{Y}^i$ is obtained as follows:

\begin{equation}
\label{eq:majorityLabel}
Y_c^i = \left\{
\begin{array}{ll}
1, & \mbox{if}\hspace{2mm} c =  \underset{{k}}{\arg \max} {N_k^i},\\
0, & \mbox{otherwise},
\end{array}
\right. \\
\end{equation} 
where $Y_c^i$ is corresponding to class $c$.



\subsection{Counting Network for Learning from Majority Label\label{sec:counting_network}}
\begin{figure}[t]
\centering
\includegraphics[width=0.86\columnwidth]{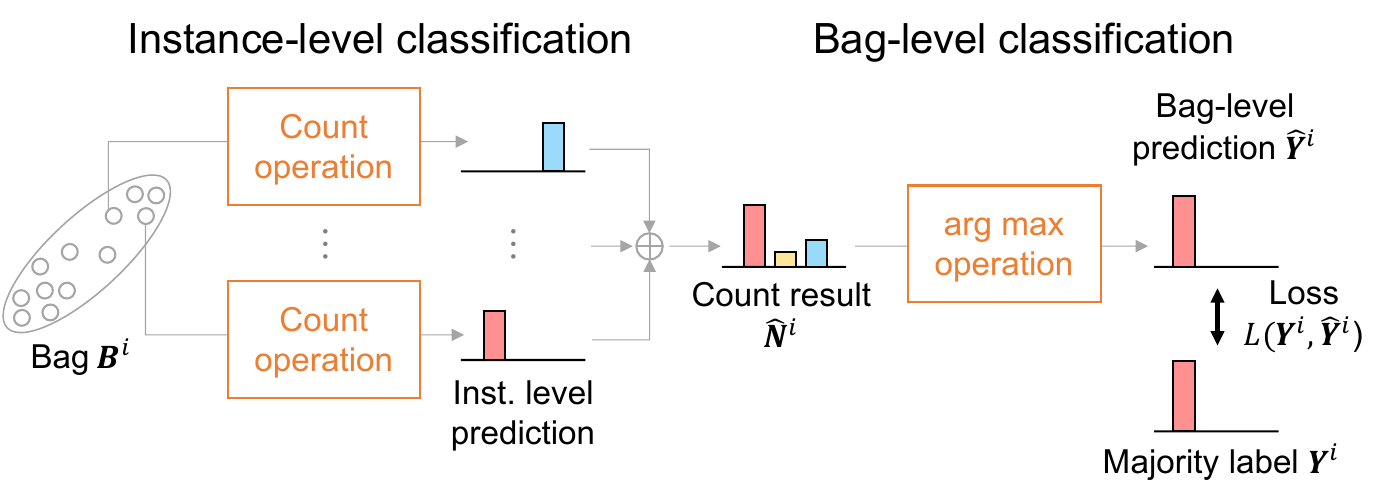}
\vspace{-3mm}
\caption{Overview of the Counting Network, which consists of two steps: instance-level classification and bag-level classifications. This model is trained so that the estimated majority label by counting the class labels of instances should be the same as the given bag-level majority label.}
\label{fig:overview}
\end{figure}

Figure~\ref{fig:overview} shows an overview of the proposed Counting Network, which consists of two steps: instance-level classification with a count operation and bag-level classification to obtain the majority class based on the counting results.
Counting Network is based on the relationship between instance-level and bag-level labels. This network estimates the majority class by counting the predicted classes for each instance and selecting the one that has the maximum value.
Then, instance-level classification is trained by calculating the bag-level loss.

To solve LML, class counting is necessary, but conventional methods did not perform this operation.
The conventional MIL method estimates bag-level labels by aggregating the class confidences for each instance through summation. However, as shown in Figure~\ref{fig:counting}(a), this approach leads to an inconsistency between the majority class obtained from the network's output and the majority class obtained from counting the predicted classes of the instances.
To avoid the inconsistency problem between the directly estimated class and the aggregation of instance classes in conventional MIL methods, the proposed Counting Network estimates the bag-level label by counting the estimated classes of the instances.



The Counting Network is trained by calculating the bag-level loss between the bag-level predictions and the majority label. 
While conventional MIL methods obtain bag-level predictions as class confidence, the proposed Counting Network obtains the bag-level prediction by performing an $\arg\max$ operation on the counting results, identifying the class with the highest number of instances.



\noindent
{\bf Instance-Level Classification:} 
In this step, the outputs for each instance are obtained through an instance-level classifier, and then they are transformed into a one-hot vector through a count operation.
Given a bag $\mathcal{B}^i = \{\bm{x}_j^i\}_{j=1}^{|\mathcal{B}^i|}$, instance-level classifier $g: \mathbb{R}^{w \times h \times d}\rightarrow [0,1]^{C}$ estimates the class of each instance within a bag.
The classifier $g$ consists of network $f : \mathbb{R}^{w \times h \times d}\rightarrow \mathbb{R}^{C}$ and softmax function with temperature $s: \mathbb{R}^{C}\rightarrow [0,1]^{C}$.
In contrast to the conventional MIL method, which uses the standard softmax function to obtain class confidence for instance-level predictions, the proposed Counting Network employs softmax with temperature to perform a count operation.
Softmax with temperature can control the distribution of the network's output vector in a differentiable manner.
By setting the temperature parameter $T$ to a small value, the largest element in the vector approaches 1, while the other elements approach 0. In other words, the element corresponding to the estimated class in the vector approaches 1, while those for the other classes approach 0.
Given a vector $\bm{z} \in \mathbb{R}^C$, this function outputs the normalized vector as follows:
\begin{equation}
\label{eq:softmax1}
\begin{split}
&s(\bm{z},T) = (s(\bm{z},T)_1,...,s(\bm{z},T)_k,...,s(\bm{z},T)_C), \\
&s(\bm{z},T)_k=\frac{\exp(\frac{z_{k}}{T})}{\sum\limits_{l=1}^C \exp(\frac{z_{l}}{T})},
\end{split}
\end{equation}
where $s(\bm{z},T)_k$ is the $k$-th element (class $k$) of output vector $s(\bm{z},T)$.
When the temperature parameter 
$T$ is set to a small value, the network output for each instance represents a pseudo-one-hot vector, i.e., only a single class value approximately takes 1, while the others approach 0.

\noindent
{\bf Bag-Level Classification by Counting Instance-Level Labels:} 
In this step, the estimated classes for the instances are counted, and the bag-level prediction is obtained by applying the $\arg\max$ operation to the count results to determine the class with the maximum value.
Given a set of estimation results for the instances,
$\{g(\bm{x}_j^i,T)\}_{j=1}^{|\mathcal{B}^i|}$, this step performs a summation operation for the estimated vector as follows:
\begin{equation}
\label{eq:sum}
\hat{N^i_k}=\sum\limits_{j=1}^{|\mathcal{B}^i|} g(\bm{x}_j^i,T)_k,
\end{equation}
since $g(\bm{x}_j^i,T)_k$ is a pseudo-one-hot vector that takes values of 0 or 1, the summation operation means counting the number of instances for each class, and $\hat{N^i_k}$ represents the number of instances belonging to class $k$.

The majority class $c$ in the bag is the class where $\hat{N^i_c}$ is the largest among all classes, and this is obtained through the $\arg\max$ operation (i.e., 
$c = \underset{{k}}{\arg \max} \hat{N^i_k}$).
Suppose we use the standard softmax as an $\arg\max$ operation to obtain the class confidence at the bag level from the count results.
In that case, the bag-level loss continues to enforce a penalty until all instances in the bag are classified as belonging to the majority class, even after the predicted bag-level class matches the majority class.
This suggests that the standard softmax can lead to overestimation of the majority class.
To avoid the overestimation of the majority class, we also use the softmax with temperature 
$s(\bm{N}^i, T)$ to obtain the majority class in a differentiable manner.
The output of this process is the one-hot vector $\hat{\bm{Y}}^i = s(\bm{N}^i, T)$, where each element of $\hat{\bm{Y}}^i$ takes a value of approximately 0 or 1, and when it is 1, it indicates the majority class of the bag.
Therefore, when the estimated bag-level class matches the ground truth majority label, the loss does not enforce a penalty, which helps suppress the overestimation of the majority class.

To train the network $g$, we use a cross-entropy loss between the estimated majority class $\hat{\bm{Y}}^i$
and its ground truth $\bm{Y}^i$ for each bag as:
\begin{equation}
\label{eq:loss}
L(\bm{Y}^i, \hat{\bm{Y}}^i) = -\sum\limits_{k=1}^C {Y}_{k}^i \log \hat{Y_{k}^i},
\end{equation}
where this loss is calculated for all the bags in the training set.
We expect that by using multiple bags and bag-level supervised labels in the training process, the network will learn to acquire the representational ability to classify each instance.
In inference, the estimated class of an instance is obtained by the classifier $g$.

\subsection{Majority Proportion Enhancement Module (MPEM)\label{sec:method_MPEM}}
In the analysis experiments of the Counting Network shown in
Figure~\ref{fig:Preliminary_analysis}
, it was revealed that in LML, instance-level classification learning is facilitated in the scenario where 
the proportion of the majority class in bags is high.
Based on this observation, we propose the Majority Proportion Enhancement Module (MPEM)\footnote{MPEM aims to facilitate learning by removing instances of the minority class from bags in the training data, and no removal is performed on the test data during inference.}, which aims to facilitate instance-level classification learning by increasing the proportion of the majority class within a bag through the removal of instances from the minority class.


In the proposed MPEM, minority class instances are removed based on instance-level predictions from the Counting Network, which was pre-trained as described in section~\ref{sec:counting_network}.
Here,  \emph{minority class} is defined as any predicted instance label that differs from the ground-truth majority class $c$ (i.e., $\hat{y}^i_j \neq c$) for a bag whose ground-truth majority class is $c$ (i.e., $Y_c^i = 1$).
However, since the model's predictions may contain errors, removing all instances predicted as a minority class could mistakenly delete majority class instances. 
To avoid this, we selectively remove instances predicted as belonging to the minority class that are distant from the majority class prototype in the feature space, as they are less likely to belong to the majority class. 
This operation reduces the risk of excessive removal and helps preserve the majority class.

\begin{figure}[t]
\centering
\includegraphics[width=0.85\columnwidth]{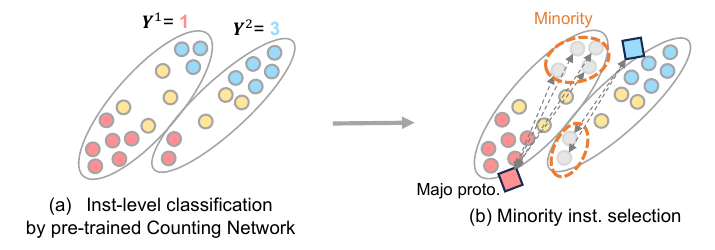}
\vspace{-2mm}
\caption{Overview of the majority proportion enhancement module. (a) Pre-trained Counting Network estimates the class of instances in the bags.
Next, for instances predicted as the minority class, the distance from the majority class prototypes is calculated, and distant instances are removed, resulting in bags with an increased proportion of the majority class.}
\label{fig:MPEM_overview}
\end{figure}

Figure~\ref{fig:MPEM_overview} shows the overview of the proposed MPEM.
First, the pre-trained Counting Network estimates the class of instances in all the bags of the training data.
Next, to obtain the feature prototype for each majority class, we collect the instances predicted to belong to class $c$ from all bags labeled with class $c$ (i.e., $Y_c^i = 1$) as follows:
$\mathcal{M}_c = \bigcup_{i=1}^n \{ \bm{x}^i_j \in \mathcal{B}^i \mid 
\hat{y}^i_j=c , \, Y_c^i = 1 \}$. 
Here, $\hat{y}^i_j$ denotes the predicted class for the instance $\bm{x}^i_j$, and $n$ is the number of bags in the training data.
The feature prototype of the majority class 
$c$ is obtained by performing a mean operation
$\bm{p}_c = \frac{1}{|\mathcal{M}_c|} \sum_{\bm{x}_j^i \in \mathcal{M}_c} \tilde{f}(\bm{x}^i_j)$, where $\tilde{f}$ is the feature extractor of network $f$.
The distance between the feature prototype $\bm{p}_c$ of the majority class $c$ and the features of the predicted minority class instances within the $i$-th bag $\mathcal{B}^i$ labeled with majority class $c$  is calculated as follows:
\begin{equation}
d_{j}^{i} =  \|\bm{p}_c - \tilde{f}(\bm{x}^i_j)\|_2, \quad \text{if} \ \hat{y}^i_j \neq  c. \,
\end{equation}
When the distance $d_{j}^{i}$ is large, the instance $\bm{x}_{j}^{i}$ is likely to belong to the minority class.
Therefore, among the instances estimated to belong to the minority class within the bag $\mathcal{B}^i$, the top $r$ ratio of those with the largest distances are removed.
Finally, the Counting Network will be retrained from initial network parameters using the bag with an enhanced majority class proportion by MPEM. This is expected to facilitate instance-level classification learning.
Here, the optimal ratio $r$ is automatically determined based on the validation loss at the best epoch of the retrained Counting Network.

\section{Experiments}
\subsection{Datasets and Experimental Setting}\label{sec:exp_setting} 

\noindent
{\bf Preparation of Bags:} 
In our experiments, we artificially created bags using four datasets.
1) CIFAR10~\cite{krizhevsky2009learning} and 2) SVHN~\cite{netzer2011reading}, both of which are commonly used in classification tasks and contain images from 10 classes.
3) PATHMNIST, a colorectal cancer histological image dataset containing 9 classes.
4) OCTMNIST, an optical coherence tomography (OCT) image dataset for retinal diseases containing 4 classes.
3) and 4) are datasets from MedMNIST~\cite{yang2023medmnist}, selected for being sufficiently large to create bags.

To create the bag, it is necessary to determine the proportion of majority class instances within the bag.
Even when the majority label for a bag is the same, there may be cases where the proportion of majority class instances within the bag differs.
In extreme cases, a class can become the majority in the bag even if its number of instances differs by only one from the second most frequent class or if all instances in the bag belong to that majority class.
In the real application, diverse proportions of the majority class, ranging from low to high, are expected in many cases.
Therefore, the goal of this study is to create a classifier model that performs effectively in the scenario {\bf Various}, which represents a setting close to the real world, where bags have various majority proportions.
In the {\bf Various} scenario, bags with majority class proportions ranging from greater than 1/$C$ to 1 were created, where $C$ is the number of classes.
The class proportions were randomly assigned, and instances were randomly sampled from each class in the labeled dataset according to these proportions to create a bag.

\noindent
{\bf Implementation Detail and Performance Metric:} 
We implemented our method using PyTorch \cite{Paszke2019PyTorchAI}.
The network model $f$ was ResNet18 \cite{he2016deep}, with weights randomly initialized. 
The network was trained using the Adam optimizer \cite{adam}, with a learning rate of $3e-4$, $\mathrm{epoch}=1500$, a mini-batch size$=64$. 
The temperature parameter $T$ is set to 0.1.
The performance of each method is evaluated with 5-fold cross-validation, and the performance of each fold is based on the epoch with the minimum validation loss.
We evaluated our model performance using instance-level classification accuracy.

\subsection{Preliminary Analysis: The Impact of the Proportion of the Majority Class on the Learning of Instance-Level Classification\label{sec:preliminary_experiment}}
LML is a novel problem setting, and it is not yet clear under what conditions instance-level classification learning becomes easier or more challenging. Among the factors that significantly influence learning in LML, the proportion of the majority class is considered one of the most critical.
Therefore, we conducted preliminary experiments in two extreme scenarios: 1) {\bf Large}: when the proportions of the majority class within the bags are randomly selected from a range of higher values (0.6 to 1) and assigned to each bag; 2) {\bf Small}: when the proportions of the majority class within the bags are randomly selected from a range of lower values (greater than 1/$C$ and up to 0.4) and assigned to each bag, where $C$ is the number of classes.
In the two scenarios, we analyzed the performance of the proposed ``Counting Network.''


Figure~\ref{fig:Preliminary_analysis} shows the instance-level classification accuracy in two scenarios across four datasets.
This result shows that the accuracy in the {\bf Large} scenario is significantly higher than in the {\bf Small} scenario.
This indicates that in LML, the proportion of the majority class within a bag is a crucial factor influencing the difficulty of instance-level classification learning and that learning becomes easier when there are more bags with a high proportion of the majority class.
When the proportion of the majority class within a bag is low, there is a high risk of minority instances being misclassified as majority instances. In contrast, when the proportion of the majority class is high, this risk is reduced, thereby enhancing the generalization ability of the classification model.


Based on this observation, we considered that instance-level classification learning can be facilitated by increasing the proportion of the majority class in bags with a low majority class proportion.
In the following section, we analyze the experimental results using the proposed method, which introduces the MPEM, a module designed to further facilitate the classification learning of the Counting Network in LML by increasing the proportion of the majority class within the bag.

\subsection{Comparison with Conventional MIL Methods\label{sec:comparison}}
To demonstrate the effectiveness of the proposed method, which includes MPEM, and the challenges faced by conventional MIL methods in solving LML, we compared our method with eight conventional MIL methods, including state-of-the-art methods.
1) ``Output+Mean''~\cite{ wang2018revisiting} aggregates the class confidences estimated for instances within a bag using averaging operations and computes the loss at the bag level.
In the MIL field, feature aggregation methods are more mainstream than output aggregation, and thus, we compared various types of 
feature aggregation methods.
2) to 5) ``Feature + Mean''~\cite{Ilse2020DeepMI}, ``Max''~\cite{Ilse2020DeepMI},`` P-norm'', ``LSE''~\cite{Pinheiro2015} aggregate instance feature in a bag using pooling operations. 
6) ``Feature + Attention"~\cite{ilse2018attention} aggregates instance features using a weighted summation, where the attention weight is estimated for each instance.
7) ``AdditiveMIL''~\cite{javed2022additive}, 8) ``AdditiveTransMIL''~\cite{javed2022additive} and 
9) ``MILLET''~\cite{early2024inherently} are the state-of-the-art methods for conventional MIL tasks. 
``AdditiveMIL'' and ``AdditiveTransMIL'' estimate the class score for each instance after multiplying the instance features with the attention weight, and then aggregating the 
class score.  Instances that contribute to the estimation of the bag-level label have high 
class scores.
``MILLET'' estimates the confidence for each instance and then performs a weighted summation of the confidences using attention weights.

Table~\ref{tab:comparison} shows the instance-level accuracies of the comparative methods in all scenarios.
``Output+mean'' estimates instance-level classes but does not consider the relationship between instance-level and bag-level labels and cannot perform counting, leading to overfitting to the bag-level label during training.
``Feature + Mean'', ``Max'', ``P-norm'', and ``LSE'' train classifiers for bag-level features, but they are unable to obtain an appropriate classifier for making predictions at the instance level, resulting in poor performance.
``Feature + attention'' assigns weights to crucial instances within the bag and performs weighted summation, but since LML determines the bag label based on the number of instances of each class in the bag, applying weights makes it difficult to identify the class with the largest count.
``AdditiveMIL'' and ``AdditiveTransMIL'' estimate instance-level class scores and perform aggregation.
However, because they assign attention weights to instance features and compute high class scores for instances that contribute to the prediction, they are unable to perform counting, which results in insufficient performance.
``MILLET'' performs a weighted summation of the confidence estimated for each instance using attention weights, making it unable to perform counting and resulting in insufficient performance.
Our method outperformed conventional MIL methods across all datasets and scenarios by being designed to consider the relationship between bag-level and instance-level labels in LML.

Furthermore, since our task is a special case of weakly supervised learning, in which instance classes are inferred from training data of bag-level labels, we added two more baselines to the comparison: the transformer-based weakly supervised method ``MCTformer+"~\cite{xu2024mctformer+} and the pseudo-labeling method ``WENO"~\cite{qu2022bidirectional}. ``MCTformer+" treats the pixel-label as an instance-label and the image-label as a bag-label. It employs multiple class-specific tokens to predict accurate class scores during image-level multi-label classification. ``WENO" assigns pseudo-labels based on attention scores used for bag-level classification, and the pseudo-labels are used for the instance-label loss in training.
These methods showed low performance in our LML setting because they were not designed for LML.

In LML, the performance at the bag level is expected to be closely linked to that at the instance level.
To analyze the bag-level classification performance of each method, Table~\ref{tab:Bag_comparison} shows the bag-level accuracies of the conventional MIL methods across all scenarios.
This result demonstrates that the proposed method is also highly effective for bag-level classification, as it consistently outperforms all comparative methods across all datasets under the challenging {\bf Small} and {\bf Various} scenarios.
The {\bf Large} scenarios represent relatively easy settings for bag-level classification, where many comparative methods achieve high performance, and the proposed method also demonstrates comparable performance.

\begin{table}[H]
    \def\@captype{table}
      \makeatother
        \centering
        \caption{Instance-level accuracies of the comparative methods on four datasets. Best performances are bold.}
        \scalebox{0.69}[0.69]{
        \begin{tabular}{c||cccc|c} 
        \hline
        \multicolumn{6} {c}{Various}\\
        \cline{0-5}
        Method &  CIFAR10 & SVHN & PATH & OCT & Average\\ \hline \hline
        Output+Mean~\cite{wang2018revisiting}  & 0.371$\pm$0.024 & 0.528$\pm$0.024  & 0.660$\pm$0.023  & 0.552$\pm$0.018 & 0.422  \\
        Feature+Mean~\cite{Ilse2020DeepMI} & 0.400$\pm$0.006  & 0.552$\pm$0.022  & 0.582$\pm$0.018  & 0.538$\pm$0.026 & 0.518 \\
        Feature+Max~\cite{Ilse2020DeepMI} & 0.293$\pm$0.014  & 0.202$\pm$0.012  & 0.508$\pm$0.025  & 0.337$\pm$0.017 & 0.335\\
        Feature+P-norm~\cite{Ilse2020DeepMI} & 0.351$\pm$0.011  &  0.289$\pm$0.012  & 0.578$\pm$0.012  & 0.448$\pm$0.015 & 0.417 \\
        Feature+LSE~\cite{Pinheiro2015} & 0.294$\pm$0.010 & 0.180$\pm$0.006  & 0.469$\pm$0.023  & 0.331$\pm$0.023 & 0.319 \\
        Feature+Attention~\cite{ilse2018attention} & 0.369$\pm$0.009 & 0.353$\pm$0.014  & 0.514$\pm$0.047  & 0.494$\pm$0.016 & 0.433 \\
        AdditiveMIL~\cite{javed2022additive} & 0.247$\pm$0.057  & 0.247$\pm$0.062  & 0.475$\pm$0.025  & 0.334$\pm$0.098 & 0.326 \\
        AdditiveTransMIL~\cite{javed2022additive} & 0.346$\pm$0.009  & 0.306$\pm$0.011  & 0.555$\pm$0.012  & 0.463$\pm$0.009 & 0.418 \\ 
        MILLET~\cite{early2024inherently} & 0.417$\pm$0.007  & 0.560$\pm$0.020 & 0.570$\pm$0.029 & 0.546$\pm$0.008 & 0.523\\
        MCTformer+~\cite{xu2024mctformer+} &0.220$\pm$0.019&0.237$\pm$0.017&0.264$\pm$0.040&0.305$\pm$0.034& 0.257\\
        WENO~\cite{qu2022bidirectional} &0.337$\pm$0.019&0.393$\pm$0.064&0.471$\pm$0.036&0.371$\pm$0.074&0.393\\
        \rowcolor{gray!15} Ours   &  \textbf{0.492}$\pm$0.013  & \textbf{0.743} $\pm$0.011 & \textbf{0.699}$\pm$0.020  & \textbf{0.634}$\pm$0.005 & \textbf{0.642} \\
        \hline
        \multicolumn{6} {c}{Small}\\
        \cline{0-5}
        Method &  CIFAR10 & SVHN & PATH & OCT & Average\\ \hline \hline
        Output+Mean~\cite{wang2018revisiting}  & 0.178$\pm$0.033 & 0.172$\pm$0.023 & 0.284$\pm$0.106 & 0.279$\pm$0.023 & 0.228  \\
        Feature+Mean~\cite{Ilse2020DeepMI} & 0.297$\pm$0.020 & 0.395$\pm$0.015 & 0.462$\pm$0.018 & 0.265$\pm$0.035 & 0.355 \\
        Feature+Max~\cite{Ilse2020DeepMI} & 0.178$\pm$0.047 & 0.108$\pm$0.008  & 0.367$\pm$0.029 & 0.256$\pm$0.018 & 0.227 \\
        Feature+P-norm~\cite{Ilse2020DeepMI} & 0.268$\pm$0.004 &  0.172$\pm$0.004 & 0.487$\pm$0.019 & 0.287$\pm$0.036 & 0.304 \\
        Feature+LSE~\cite{Pinheiro2015} & 0.204$\pm$0.010 & 0.120$\pm$0.006 & 0.374$\pm$0.025 & 0.268$\pm$0.008 & 0.242 \\
        Feature+Attention~\cite{ilse2018attention} & 0.163$\pm$0.011 & 0.108$\pm$0.010 & 0.350$\pm$0.146 & 0.280$\pm$0.034 & 0.225 \\
        AdditiveMIL~\cite{javed2022additive} & 0.147$\pm$0.033 & 0.110$\pm$0.006 & 0.254$\pm$0.064 & 0.255$\pm$0.008 & 0.192 \\
        AdditiveTransMIL~\cite{javed2022additive} & 0.270$\pm$0.010 & 0.126$\pm$0.028 & 0.488$\pm$0.014 & 0.306$\pm$0.008 & 0.298 \\ 
        MILLET~\cite{early2024inherently} &0.303$\pm$0.010&0.395$\pm$0.010&0.452$\pm$0.014&0.269$\pm$0.038&0.355\\
        MCTformer+~\cite{xu2024mctformer+} &0.113$\pm$0.020&0.121$\pm$0.012&0.097$\pm$0.015&0.238$\pm$0.018&0.142 \\
        WENO~\cite{qu2022bidirectional} &0.164$\pm$0.048&0.198$\pm$0.039&0.245$\pm$0.116&0.244$\pm$0.010&0.213\\
        \rowcolor{gray!15} Ours &\textbf{0.368}$\pm$0.016&\textbf{0.604}$\pm$0.021&\textbf{0.612}$\pm$0.017&\textbf{0.421}$\pm$0.024&\textbf{0.501} \\
        \hline
        \multicolumn{6} {c}{Large}\\
        \cline{0-5}
        Method &  CIFAR10 & SVHN & PATH & OCT & Average\\ \hline \hline
        Output+Mean~\cite{wang2018revisiting}  & 0.532$¥\pm$0.012 & 0.808$¥\pm$0.016 & 0.732$¥\pm$0.012 & 0.675$¥\pm$0.007 & 0.687 \\
        Feature+Mean~\cite{Ilse2020DeepMI} & 0.460$\pm$0.022  & 0.667$\pm$0.022 & 0.616$\pm$0.011 & 0.656$\pm$0.007 & 0.600 \\
        Feature+Max~\cite{Ilse2020DeepMI} & 0.340$\pm$0.017 & 0.276$\pm$0.012 & 0.512$\pm$0.015 & 0.370$\pm$0.017 & 0.375\\
        Feature+P-norm~\cite{Ilse2020DeepMI} & 0.402$\pm$0.016 & 0.399$\pm$0.013 & 0.568$\pm$0.020 & 0.490$\pm$0.004 & 0.465\\
        Feature+LSE~\cite{Pinheiro2015} & 0.341$\pm$0.028 & 0.229$\pm$0.008 & 0.518$\pm$0.018 & 0.346$\pm$0.017 & 0.359\\
        Feature+Attention~\cite{ilse2018attention} & 0.406$\pm$0.016 & 0.500$\pm$0.024 & 0.591$\pm$0.015 & 0.561$\pm$0.012 & 0.515 \\
        AdditiveMIL~\cite{javed2022additive} & 0.406$\pm$0.019 & 0.524$\pm$0.056 & 0.599$\pm$0.029 & 0.411$\pm$0.131 & 0.485 \\
        AdditiveTransMIL~\cite{javed2022additive} & 0.390$\pm$0.010 & 0.478$\pm$0.038 & 0.520$\pm$0.017 & 0.536$\pm$0.007& 0.481 \\ MILLET~\cite{early2024inherently} &0.476$\pm$0.029& 0.647$\pm$0.0263 & 0.601$\pm$ 0.012&0.656$\pm$0.009&0.595\\ 
        MCTformer+~\cite{xu2024mctformer+} &0.329$\pm$0.041&0.345$\pm$0.042&0.334$\pm$0.049&0.387$\pm$0.063& 0.349\\
        WENO~\cite{qu2022bidirectional} &0.292$\pm$0.039&0.389$\pm$0.045&0.285$\pm$0.066&0.405$\pm$0.033&0.343\\
        \rowcolor{gray!15} Ours   & \textbf{0.616}$\pm$0.009  & \textbf{0.851}$\pm$0.006 & \textbf{0.812}$\pm$0.007 & \textbf{0.714}$\pm$0.009 & \textbf{0.748} \\
        \hline
        \end{tabular}
        }
        \label{tab:comparison}
\end{table}

\begin{table}[H]
    \def\@captype{table}
      \makeatother
        \centering
        \caption{Bag-level accuracies of the comparative methods on four datasets. Best performances are bold.}
        \scalebox{0.75}[0.75]{
        \begin{tabular}{c||cccc|c} 
        \hline
        \multicolumn{6} {c}{Various}\\
        \cline{0-5}
        Method &  CIFAR10 & SVHN & PATH & OCT & Average\\ \hline \hline
        Output+Mean~\cite{wang2018revisiting}  &0.514$\pm$0.093&0.802$\pm$0.019&0.816$\pm$0.050&0.685$\pm$0.050&0.704\\
        Feature+Mean~\cite{Ilse2020DeepMI} &0.764$\pm$0.033&0.816$\pm$0.016&0.804$\pm$0.022&0.850$\pm$0.028&0.809\\
        Feature+Max~\cite{Ilse2020DeepMI} &0.470$\pm$0.050&0.340$\pm$0.044&0.592$\pm$0.036&0.529$\pm$0.060&0.483\\
        Feature+P-norm~\cite{Ilse2020DeepMI} &0.648$\pm$0.047&0.578$\pm$0.060&0.790$\pm$0.024&0.729$\pm$0.041& 0.686\\
        Feature+LSE~\cite{Pinheiro2015} &0.418$\pm$0.066&0.388$\pm$0.036&0.668$\pm$0.044&0.545$\pm$0.074&0.505\\
        Feature+Attention~\cite{ilse2018attention} &0.660$\pm$0.075&0.568$\pm$0.025&0.740$\pm$0.046&0.721$\pm$0.034&0.672\\
        AdditiveMIL~\cite{javed2022additive} &0.444$\pm$0.158&0.442$\pm$0.187&0.724$\pm$0.053&0.453$\pm$0.183&0.516\\
        AdditiveTransMIL~\cite{javed2022additive} & 0.732$\pm$0.047&0.648$\pm$0.024&0.816$\pm$0.034&0.773$\pm$0.018&0.742\\ 
        MILLET~\cite{early2024inherently} &0.784$\pm$0.024&0.836$\pm$0.015&0.846$\pm$0.019&0.818$\pm$0.036&0.821\\
        \rowcolor{gray!15} Ours   &\textbf{0.790}$\pm$0.043&\textbf{0.904}$\pm$0.015&\textbf{0.850}$\pm$0.021&\textbf{0.830}$\pm$0.032& \textbf{0.844}\\
        \hline
        \multicolumn{6} {c}{Small}\\
        \cline{0-5}
        Method &  CIFAR10 & SVHN & PATH & OCT & Average\\ \hline \hline
        Output+Mean~\cite{wang2018revisiting}  &0.134$\pm$0.023&0.150$\pm$0.070&0.204$\pm$0.052&0.344$\pm$0.029&0.208\\
        Feature+Mean~\cite{Ilse2020DeepMI} &0.350$\pm$0.055&0.416$\pm$0.054&0.410$\pm$0.062&0.280$\pm$0.084&0.364\\
        Feature+Max~\cite{Ilse2020DeepMI} &0.136$\pm$0.026&0.106$\pm$0.058&0.230$\pm$0.040&0.230$\pm$0.060&0.176\\
        Feature+P-norm~\cite{Ilse2020DeepMI} &0.242$\pm$0.040&0.194$\pm$0.033&0.318$\pm$0.057&0.242$\pm$0.072&0.249\\
        Feature+LSE~\cite{Pinheiro2015} &0.176$\pm$0.026&0.096$\pm$0.042&0.222$\pm$0.039&0.262$\pm$0.023&0.164\\
        Feature+Attention~\cite{ilse2018attention} &0.146$\pm$0.041&0.108$\pm$0.025&0.238$\pm$0.101&0.302$\pm$0.059&0.199\\
        AdditiveMIL~\cite{javed2022additive} &0.150$\pm$0.031&0.128$\pm$0.033&0.220$\pm$0.063&0.336$\pm$0.052&0.209\\
        AdditiveTransMIL~\cite{javed2022additive} &0.250$\pm$0.035&0.148$\pm$0.028&0.464$\pm$0.034&0.342$\pm$0.021&0.301\\ 
        MILLET~\cite{early2024inherently} &0.342$\pm$0.030&0.440$\pm$0.062&0.390$\pm$0.052&0.298$\pm$0.043&0.368\\
        \rowcolor{gray!15} Ours &\textbf{0.362}$\pm$0.068&\textbf{0.534}$\pm$0.030&\textbf{0.498}$\pm$0.052&\textbf{0.406}$\pm$0.059&\textbf{0.450}\\
        \hline
        \multicolumn{6} {c}{Large}\\
        \cline{0-5}
        Method &  CIFAR10 & SVHN & PATH & OCT & Average\\ \hline \hline
        Output+Mean~\cite{wang2018revisiting} &0.978$\pm$0.029&\textbf{1.000}$\pm$0.000&0.996$\pm$0.005&0.958$\pm$0.025&0.983\\
        Feature+Mean~\cite{Ilse2020DeepMI} &\textbf{0.998}$\pm$0.004&\textbf{1.000}$\pm$0.000&\textbf{1.000}$\pm$0.000&0.994$\pm$0.008&\textbf{0.998} \\
        Feature+Max~\cite{Ilse2020DeepMI} &0.804$\pm$0.062&0.732$\pm$0.047&0.926$\pm$0.014&0.750$\pm$0.028&0.803\\
        Feature+P-norm~\cite{Ilse2020DeepMI} &0.960$\pm$0.014&0.950$\pm$0.021&0.992$\pm$0.007&0.932$\pm$0.010&0.959\\
        Feature+LSE~\cite{Pinheiro2015} &0.868$\pm$0.039&0.752$\pm$0.057&0.944$\pm$0.024&0.780$\pm$0.052&0.836\\
        Feature+Attention~\cite{ilse2018attention} &0.988$\pm$0.007&0.998$\pm$0.004&\textbf{1.000}$\pm$0.000&0.976$\pm$0.024&0.990\\
        AdditiveMIL~\cite{javed2022additive} &0.974$\pm$0.010&0.998$\pm$0.004&\textbf{1.000}$\pm$0.000&0.762$\pm$0.216&0.934\\
        AdditiveTransMIL~\cite{javed2022additive} &0.984$\pm$0.008&\textbf{1.000}$\pm$0.000&0.996$\pm$0.005&0.994$\pm$0.005&0.994\\
        MILLET~\cite{early2024inherently} &0.996$\pm$0.005&\textbf{1.000}$\pm$0.000&0.950$\pm$0.029&0.994$\pm$0.049&0.985\\ 
        \rowcolor{gray!15} Ours   &0.992$\pm$0.012&\textbf{1.000}$\pm$0.000&0.998$\pm$0.004&\textbf{0.996}$\pm$0.005&0.997\\
        \hline
        \end{tabular}
        }
        \label{tab:Bag_comparison}
\end{table}


\begin{table}[t]
    \def\@captype{table}
      \makeatother
        \centering
        \caption{Ablation study. Instance-level accuracies of the methods that use each operation on four datasets.  Best performances are bold.}
        \scalebox{0.55}[0.55]{
        \begin{tabular}{c|ccc||cccc|c} 
        \hline
         \multicolumn{9}{c}{various} \\  \hline
        Method & $\arg\max$ & Count & MPEM  &  CIFAR10 & SVHN & PATH & OCT & Average \\ \hline \hline
        Output+Mean~\cite{wang2018revisiting}  &&&& 0.371$\pm$0.024 & 0.528$\pm$0.024  & 0.660$\pm$0.023  & 0.552$\pm$0.018 & 0.422  \\
        Counting Network w/o Count &$\checkmark$&&& 0.467$\pm$0.003  & 0.718$\pm$0.007  & 0.655$\pm$0.022  & 0.593$\pm$0.035 & 0.608  \\
        Counting Network  &$\checkmark$& $\checkmark$ &&  0.487$\pm$0.012  & 
        0.721$\pm$0.011  & 0.675$\pm$0.017  & 0.627$\pm$0.011 & 0.628 \\ 
        Ours & $\checkmark$ &$\checkmark$& $\checkmark$ &\textbf{0.492}$\pm$0.013  & \textbf{0.743} $\pm$0.011 & \textbf{0.699}$\pm$0.020  & \textbf{0.634}$\pm$0.005 & \textbf{0.642}\\
        \hline
         \multicolumn{9}{c}{Small} \\  \hline
        Method & $\arg\max$ & Count & MPEM  &  CIFAR10 & SVHN & PATH & OCT & Average \\ \hline \hline
        Output+Mean~\cite{wang2018revisiting}  &&&& 0.178$\pm$0.033 & 0.172$\pm$0.023 & 0.284$\pm$0.106 & 0.279$\pm$0.023 & 0.228  \\
        Counting Network w/o Count & $\checkmark$ &&& 0.318$\pm$0.017 & 0.390$\pm$0.011 & 0.535$\pm$0.017 & 0.265$\pm$0.025  & 0.377  \\
        Counting Network   &$\checkmark$& $\checkmark$ && 0.335$\pm$0.015 & 0.479$\pm$0.051 &   0.578$\pm$0.027 & 0.392$\pm$0.036 & 0.446 \\ 
        Ours & $\checkmark$ &$\checkmark$& $\checkmark$ &\textbf{0.368}$\pm$0.016&\textbf{0.604}$\pm$0.021&\textbf{0.612}$\pm$0.017&\textbf{0.421}$\pm$0.024&\textbf{0.501}\\
        \hline
         \multicolumn{9}{c}{Large} \\  \hline
        Method & $\arg\max$ & Count & MPEM &  CIFAR10 & SVHN & PATH & OCT & Average \\ \hline \hline
        Output+Mean~\cite{wang2018revisiting}  &&&& 0.532$¥\pm$0.012 & 0.808$¥\pm$0.016 & 0.732$¥\pm$0.012 & 0.675$¥\pm$0.007 & 0.687  \\
        Counting Network w/o Count &$\checkmark$&&& 0.589$\pm$0.009 & \textbf{0.852}$\pm$0.003 & 0.782$\pm$0.007 & \textbf{0.719}$\pm$0.007 & 0.736 \\
        Counting Network  &$\checkmark$& $\checkmark$ &&  0.614$\pm$0.004  & 0.847$\pm$0.003 & 0.805$\pm$0.005 & 0.699$\pm$0.042 & 0.741\\ 
        Ours & $\checkmark$ &$\checkmark$& $\checkmark$ & \textbf{0.616}$\pm$0.009  & 0.851$\pm$0.006 & \textbf{0.812}$\pm$0.007 & 0.714$\pm$0.009 & \textbf{0.748}\\
        \hline
        \end{tabular}
        }
        \label{tab:ablation}
\end{table}

\subsection{Ablation Study\label{sec:ablation}}
To investigate the effectiveness of the count operation, the $\arg\max$ operation, and MPEM, we conduct an ablation study.
We compared our method with three ablation methods.
``Output+Mean'' is a baseline method that removes the $\arg\max$ operation, count operation, and MPEM from our method. This method aggregates the confidence of instances in a bag using the mean operation for bag-level estimation.
``Counting Network w/o Count'' removes both the count operation and MPEM from our method. Instead of using the softmax with temperature, it employs the standard softmax for instance-level classification and performs bag-level prediction by summing the confidence scores.
``Counting Network'' removes the MPEM from our method. This method utilizes the Counting Network for training but does not use MPEM, which is designed to increase the proportion of the majority class.

Table~\ref{tab:ablation} shows the accuracies of each ablation method.
Introducing the $\arg\max$ operation improved performance across all scenarios. 
``Output+Mean'' continues to impose a loss during training even when the predicted bag class matches the majority label. This persists until all instances within the bag are classified as belonging to the majority class, which may result in overestimation.
The ``Counting Network,'' which introduces the count operation, improved performance in most settings.
It significantly improved performance, particularly in the {\bf Small} and {\bf Various} scenarios. 
In the {\bf Large} scenario, there are few bags in which the proportion of the majority class is close to that of the minority class. In such cases, the difference in the sum of confidence between the majority and minority classes becomes large, resulting in a low probability of the label switching to the minority class. 
Therefore, the effectiveness of maintaining consistency through the count mechanism is reduced.
Furthermore, the proposed method improved performance by introducing MPEM. 
In particular, significant performance improvement is observed in the Small and Various settings, where many bags contain a low proportion of the majority class. This demonstrates the effectiveness of increasing the majority class proportion in such bags.


\subsection{Detail Analysis of Counting Network  \label{sec:analysis_count}} 
In this section, we present a detailed analysis of the effectiveness of each operation in the proposed ``Counting Network,'' which estimates bag-level labels by counting the instance classes and performs instance-level classification learning through bag-level loss.

\begin{figure}[t]
 \begin{center}
\includegraphics[width=0.52\columnwidth]{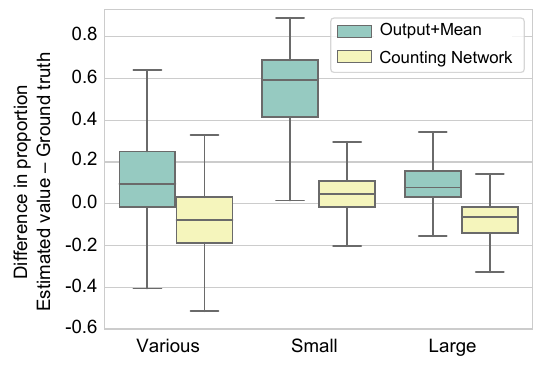}
     \vspace{-2mm}
    \caption{The majority class proportion estimation performance of the proposed ``Counting Network'' and ``Output+Mean.''
    The vertical axis represents the subtraction value (estimated majority class proportion - ground truth majority class proportion). The horizontal axis represents the names of each scenario. The subtraction value was calculated using all datasets.}
    \label{fig:MajorityProportion}
 \end{center}
     \vspace{-2mm}
\end{figure}

To verify the ability of the proposed ``Counting Network'' to estimate the majority class proportion and the overestimation of the baseline method ``Output+Mean,'' we present the difference between the estimated majority class proportion and the ground truth majority class proportion.
Figure~\ref{fig:MajorityProportion} shows a box plot representing the distribution of the difference between the estimated majority class proportion and the ground truth majority class proportion for each bag in each scenario using all datasets. The vertical axis represents the subtraction value  (estimated majority class proportion - ground truth majority class proportion). A value of 0 indicates accurate estimation, while a value greater or less than 0 indicates overestimation and underestimation of the majority proportion, respectively. The horizontal axis represents the names of each scenario.
This result shows that the baseline method led to overestimation in all scenarios. In particular, a significant difference between the proposed ``Counting Network'' and the baseline method was observed in the {\bf Small} scenario, contributing to the remarkable performance gap.

To demonstrate that the proposed ``Counting Network'' maintains consistency between the majority class predicted by the network outputs and the one obtained by counting the number of instances, we compared the consistency rate.
Here, the consistency rate is defined as the ratio of cases where the majority class obtained by counting the instances' classes matches the majority class obtained by aggregating the network outputs, to the total number of bags where the estimated majority class is correct.

\begin{table}[t]
    \def\@captype{table}
      \makeatother
        \centering
        \caption{Consistency rate between labels estimated by aggregation and by counting instances. The rate is the average on all datasets.  Best performances are bold.}
        \scalebox{0.8}[0.8]{
        \begin{tabular}{c||c|c|c} 
         \hline
          Method & \multicolumn{1}{c|}{Various} & \multicolumn{1}{c|}{Small} & \multicolumn{1}{c}{Large}\\ \hline 
        Output+Mean~\cite{wang2018revisiting}   & 0.961$\pm$0.032& 0.957$\pm$0.074 & 0.997$\pm$0.005\\
        Feature+Mean~\cite{Ilse2020DeepMI}   & 0.892$\pm$0.050& 0.690$\pm$0.176 & 0.987$\pm$0.012\\
        Feature+Max~\cite{Ilse2020DeepMI}    & 0.456$\pm$0.224& 0.523$\pm$0.395 & 0.568$\pm$0.187\\
        Feature+P-norm~\cite{Ilse2020DeepMI}   & 0.762$\pm$0.120& 0.458$\pm$0.283 & 0.925$\pm$0.039\\
        Feature+LSE~\cite{Pinheiro2015}    & 0.494$\pm$0.203 & 0.290$\pm$0.177 & 0.609$\pm$0.208\\
        Feature+Attention~\cite{ilse2018attention}   & 0.730$\pm$0.129& 0.831$\pm$0.214 & 0.911$\pm$0.077\\
        AdditiveMIL~\cite{javed2022additive}    & 0.469$\pm$0.189& 0.526$\pm$0.394 & 0.736$\pm$0.233\\
        AdditiveTransMIL~\cite{javed2022additive}    & 0.780$\pm$0.074& 0.494$\pm$0.153 & 0.889$\pm$0.089\\
        MILLET~\cite{early2024inherently} &0.906$\pm$0.046&0.710$\pm$0.151&0.983$\pm$0.025 \\
        Counting Network w/o Count   & 0.973$\pm$0.016& 0.914$\pm$0.072 & 0.999$\pm$0.003\\
         \rowcolor{gray!15} Counting Network   & \textbf{0.990}$\pm$0.011& \textbf{0.963}$\pm$0.027 & \textbf{1.00}$\pm$0.000\\
        \hline
        \end{tabular}
}
        \label{tab:consistency}
        \vspace{-3mm}
\end{table}

Table~\ref{tab:consistency} shows the consistency rates for the conventional MIL methods, ``Counting Network w/o Count,'' and proposed ``Counting Network'' in each scenario.
Here, the consistency rate is the average across all datasets.
This result indicates that the feature‑aggregation approaches frequently suffer from low consistency at inference time because the classifier receives different types of inputs at the bag and instance levels. At the bag level, the model predicts a label from a single vector obtained by aggregating the instance features, which blurs differences among instances. At the instance level, each feature is evaluated on its own, so the decision boundaries learned from aggregated data may not align with those required for individual instances. This mismatch between training signals leads to inconsistent predictions across levels.
``AdditiveMIL,'' ``AdditiveTransMIL,'' and ``MILLET'' compute the bag score as a weighted sum of the instance‑level scores, giving the highest weights to a few “informative” instances. 
Although this strategy improves bag-level accuracy, it causes a mismatch at inference time, as the influence of certain instances becomes dominant regardless of the number of instances in the bag.
``Output+Mean'' and ``Counting Network w/o Count'' aggregate the per‑instance confidence scores by averaging them, rather than counting the number of predictions for each class. A few high‑confidence minority instances can outweigh many moderate‑confidence majority ones, so the class with the highest mean confidence may differ from the most frequent class, as shown in Figure~\ref{fig:counting}, reducing consistency.

The proposed ``Counting Network'' achieved the highest consistency rate by counting the estimated classes of instances using a count operation.
Especially in the {\bf Small} and {\bf Various} scenarios, the consistency rate was significantly improved compared to the "Counting Network w/o Count."
In these settings, there are many bags where the proportion of the majority class is close to that of the minority class. In such cases, the difference in the sum of the confidence between the majority and minority classes becomes small, leading to an increased probability of the label switching to the minority class. 
By performing the count operation and representing the estimated result for the instance as a one-hot vector, it is possible to count the instances, thereby solving this issue.

\subsection{Detail Analysis of MPEM\label{sec:Analsis_MPEM}}


\begin{table}[t]
    \def\@captype{table}
      \makeatother
        \centering
        \caption{Instance-level accuracies of the proposed method on four datasets under the {\bf Various} scenario when the removal ratio $r$ of the MPEM module is changed. Best performances are bold.}
        \scalebox{0.8}[0.8]{
        \begin{tabular}{c||cccc|c} 
        \hline
        $r$-ratio  &  CIFAR10 & SVHN & PATH & OCT & Average \\ \hline \hline
        0&  0.487$\pm$0.012  & 
        0.721$\pm$0.011  & 0.675$\pm$0.017  & 0.627$\pm$0.011 & 0.628\\
        0.1&0.486$\pm$0.009&0.718$\pm$0.011&0.680$\pm$0.016&0.632$\pm$0.006& 0.629\\
        0.2&0.492$\pm$ 0.006&0.721$\pm$0.012&0.673$\pm$0.010&0.629$\pm$0.012& 0.629\\
        0.3&0.491$\pm$0.011&0.722$\pm$0.006&0.673$\pm$0.015&\textbf{0.634}$\pm$0.008& 0.630\\
        0.4&\textbf{0.504}$\pm$0.004&0.728$\pm$0.008&0.684$\pm$0.025&0.630$\pm$0.007& 0.637\\
        0.5&0.495$\pm$0.005&0.737$\pm$0.011&0.683$\pm$0.020&0.626$\pm$0.009& 0.635\\
        0.6&0.499$\pm$0.007&0.737$\pm$0.011&0.686$\pm$0.013&0.622$\pm$0.015& 0.636\\
        0.7&0.495$\pm$0.003&0.739$\pm$0.005&0.695$\pm$0.018&0.620$\pm$0.013& 0.637\\
        0.8&0.500$\pm$0.006&0.717$\pm$0.051&0.706$\pm$0.011&0.609$\pm$0.020& 0.633\\
        0.9&0.489$\pm$0.008&0.716$\pm$0.042&\textbf{0.709}$\pm$0.009&0.614$\pm$0.015& 0.632\\
        1.0&0.477$\pm$0.013&0.699$\pm$0.057&0.682$\pm$0.018&0.595$\pm$0.018& 0.613\\
        \rowcolor{gray!15}  Minimum loss (Ours)    &  0.492$\pm$0.013  & \textbf{0.743}$\pm$0.011  & 0.699$\pm$0.020  & \textbf{0.634}$\pm$0.005 & \textbf{0.642}\\
        \hline
        \end{tabular}
        }
        \label{tab:MPEM_various_k}
\end{table}

In this section, we conduct a detailed analysis to demonstrate the effectiveness of the proposed MPEM, which facilitates instance-level classification learning by increasing the proportion of the majority class within the bag.

\begin{figure}[t]
 \begin{center}
\includegraphics[width=1.05\columnwidth]{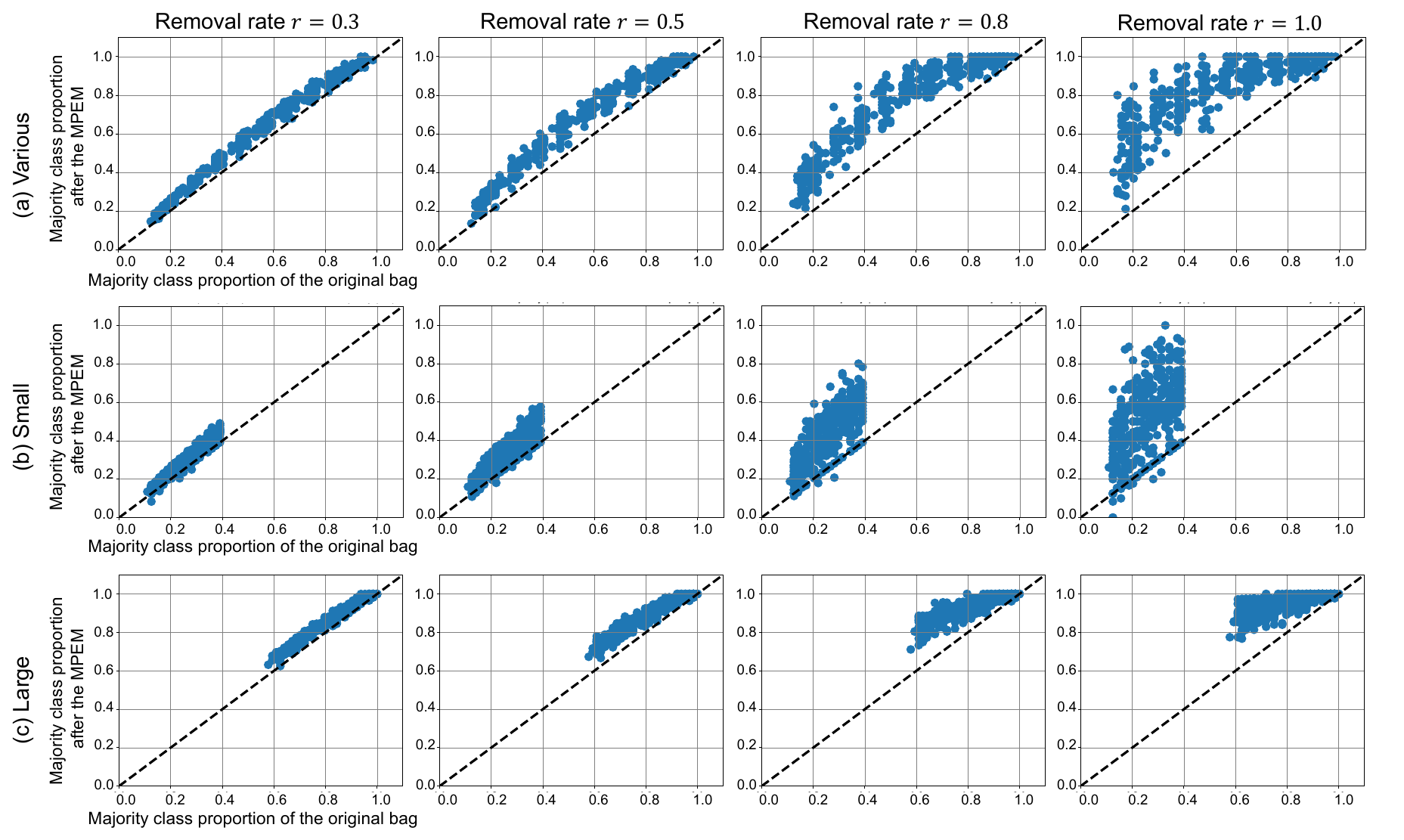}
    \caption{The relationship between the majority proportion of the original bag and the proportion after applying the MPEM when the removal ratio $r$ is changed, with each blue plot point representing a single bag. This analysis was conducted using the CIFAR10 dataset.}
    \label{fig:MajorityProportion_scatter}
 \end{center}
\end{figure}

To analyze the relationship between the removal ratio $r$ and instance-level accuracy, we evaluated the performance by changing the removal ratio $r$ under a {\bf Various} scenario of the CIFAR10 dataset.
Table~\ref{tab:MPEM_various_k} shows the instance-level accuracy for different values of the removal ratio $r$. 
The values shown above the dashed line represent the instance-level  accuracy when 
$r$ was fixed across all folds and averaged.
 On the other hand, the values shown below the dashed line represent the instance-level accuracy averaged over the folds, where $r$ was automatically selected to minimize the bag-level validation loss for each fold. 
 This result indicates that while performance improved for certain values of $r$, removing all instances estimated as belonging to the minority class ($r=1.0$) leads to mistakenly deleting majority class instances as well, resulting in a decline in performance across most datasets.
Therefore, it is important to selectively remove instances with a high probability of belonging to the minority class, and it is clear that removing all instances estimated as such is not appropriate.
Additionally, this result demonstrates that the proposed method can automatically select an approximately optimal $r$ by choosing the $r$ that minimizes the validation loss in each fold. 

To analyze the increase in the majority class proportion within each bag when the removal ratio $r$ is changed, 
Figure~\ref{fig:MajorityProportion_scatter} shows a y-y plot of the majority proportions, comparing those of the original bag (horizontal) and the bag after applying MPEM (vertical).
Each plot point corresponds to a single bag. Bags with no change in proportion are plotted along the ascending dashed line, and the further a point is above the dashed line, the greater the increase in proportion. The visualization was performed across all scenarios for removal ratios of 0.3, 0.5, 0.8, and 1.0, using the CIFAR10 dataset.
These results indicate that increasing the removal ratio successfully increases the proportion of the majority class in all scenarios.
In particular, for bags in the {\bf Small} scenario and bags with a small majority class proportion in the {\bf Various} scenario, it can be confirmed that many bags show a large increase in the proportion.
In particular, it can be observed that many bags in the {\bf Small} scenario and bags with a small majority class proportion in the {\bf Various} scenario, which are originally difficult to learn from, are transformed into bags with relatively larger majority class proportions that are more conducive to learning.
This suggests that MPEM leads to a significant improvement in accuracy in the {\bf Small} and {\bf Various} scenarios.

\begin{figure}[t]
 \begin{center}
\includegraphics[width=0.5\columnwidth]{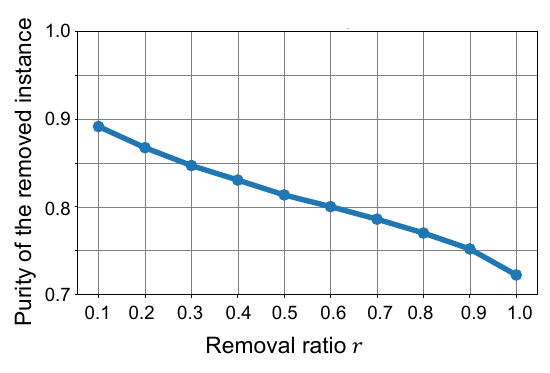}
    \caption{The purity of the removed instances when the removal ratio $r$ is changed. 
    Purity is defined as the number of minority class instances among the removed instances, divided by the total number of removed instances. This analysis was conducted using the all dataset under the {\bf Various} scenario.}
    \label{fig:removal_inst_purity}
 \end{center}
\end{figure}

Increasing the removal ratio $r$ can increase the proportion of the majority class, but if the ratio is too large, it is expected that more majority class instances will be mistakenly removed.
To analyze the relationship between the removal ratio
$r$ and the number of majority class instances mistakenly removed, 
Figure~\ref{fig:removal_inst_purity} shows the purity of the removed instances on the vertical axis and the removal ratio $r$ on the horizontal axis, with the analysis conducted using the all dataset under the {\bf Various} scenario.
Here, purity is a rate defined as the number of minority class instances among the removed instances divided by the total number of removed instances.
Therefore, when the purity takes a low value, the removed instances include a large number of majority class instances.
This result shows that as the removal ratio $r$ increases, the purity decreases.
From this observation, it can be assumed that if the removal ratio
$r$ is too large, the number of majority class instances that are mistakenly deleted increases, which leads to a decrease in performance.

From Figures~\ref{fig:MajorityProportion_scatter} and \ref{fig:removal_inst_purity}, it can be observed that while removing all instances predicted to belong to the minority class successfully increases the proportion of the majority class, setting too large a removal ratio leads to excessive removal, resulting in a decline in performance.
Therefore, there is a trade-off between the increase in the proportion of the majority class by raising the removal rate and the reduction in the number of majority instances within the bag.
From this observation, it is clear that determining the optimal removal rate 
$r$ is important. The proposed method successfully selects an approximately optimal removal ratio 
$r$ based on the validation loss.

\begin{figure}[t]
 \begin{center}
\includegraphics[width=0.5\columnwidth]{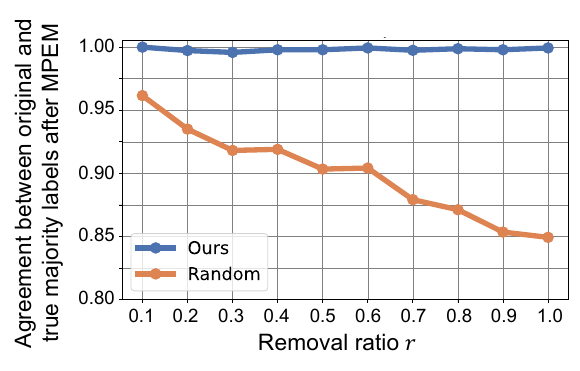}
    \caption{The agreement between the original bag labels and the true majority class in the bag after applying MPEM  when the removal ratio $r$ is changed. This analysis was conducted using the all dataset under the {\bf Various} scenario.}
    \label{fig:bag_label_agreement}
 \end{center}
\end{figure}

To evaluate the agreement between the original bag-level labels and the true majority class in the bag after applying MPEM, we present Figure~\ref{fig:bag_label_agreement}, which shows the relationship between each removal ratio $r$ and the agreement rate on the all dataset under the {\bf Various} scenario.
As a baseline, the evaluation also shows the results when the same number of instances as MPEM are removed through random sampling (``Random'').
Here, the agreement rate is defined as the number of bags in which the original bag-level labels match the true majority class in the bag after applying MPEM, divided by the total number of bags.
This result indicates that the bag-level majority label remains intact after instance removal by MPEM: the agreement between the original majority label and the post-removal label stays at 1.0 despite increasing removal ratios $r$, whereas random removal causes the agreement to drop as $r$ increases. Even if the pre-trained Counting Network misclassifies some instances, the impact on the bag-level majority class is minimal. Because the bag label does not change, learning benefits from a higher proportion of majority-class instances within each bag without introducing label noise, which ultimately leads to higher accuracy.


\section{Conclusion}
In this paper, we propose a novel multi-class Multiple-Instance Learning (MIL) problem, called Learning from Majority Label (LML), in which the majority class within a bag is assigned as the bag-level label.
The goal of LML is to train a model to classify each instance using bag-level majority class labels without relying on instance-level labels.
LML is valuable in a variety of applications, including pathology image segmentation, political voting prediction, customer sentiment analysis, and environmental monitoring.

To solve LML, we proposed the Counting Network, which is trained by counting the estimated classes of instances while ensuring that the class with the highest count matches the bag-level label.
The conventional MIL method, which aggregates instance-level class confidence by summation, is unsuitable for LML because it may cause inconsistencies between the estimated bag-level labels and the labels obtained by counting the number of instances in each class.
In contrast, the proposed Counting Network performs aggregation by counting the instances, which leads to consistency between the estimated bag-level labels and the labels obtained by counting the number of instances for each class.


Furthermore, the analysis experiments revealed that the proportion of the majority class within a bag is a crucial factor influencing instance-level classification learning.
In particular, bags with a higher proportion of the majority class were found to further facilitate learning.
Based on this observation, we developed the Majority Proportion Enhancement Module (MPEM) as a module to facilitate learning, which increases the proportion of the majority class by removing minority class instances within the bags.
Experiments demonstrated the superiority of the proposed method on four datasets compared to conventional MIL methods. Moreover, ablation studies confirmed the effectiveness of each module.


The limitation of this paper is that there is insufficient development of real-world task datasets for the application of LML. LML is a novel problem setting with various potential applications, and it is a field that will be widely studied in the future. However, existing MIL datasets are designed for binary or multi-label classification, which differs from the LML setup.
Therefore, in order to advance LML, the development of benchmark datasets is necessary.

In addition, a key future direction is to provide a theoretical analysis that supports the empirical findings presented in this paper.
Since this paper empirically demonstrates the effectiveness of MPEM, it is highly plausible that theoretical guarantees can be established under reasonable assumptions.
We plan to pursue extending analyses based on rademacher complexity and covering numbers to the LML setting as an important direction for future research.


\clearpage
\section*{Acknowledgments}
This work was supported by JST BOOST, Japan Grant Number JPMJBS2406, SIP-JPJ012425, JSPS KAKENHI Grant Number JP24K03002, JP23KJ1723, JP23K18509, and JST ACT-X JPMJAX23CR.

\bibliography{bibfile}

\begin{thebibliography}{10}
\expandafter\ifx\csname url\endcsname\relax
  \def\url#1{\texttt{#1}}\fi
\expandafter\ifx\csname urlprefix\endcsname\relax\def\urlprefix{URL }\fi
\expandafter\ifx\csname href\endcsname\relax
  \def\href#1#2{#2} \def\path#1{#1}\fi

\bibitem{ramon2000multi}
J.~Ramon, L.~De~Raedt, {Multi Instance Neural Networks}, in: International Conference on Machine Learning workshop, 2000, pp. 53--60.

\bibitem{wang2018revisiting}
X.~Wang, Y.~Yan, P.~Tang, X.~Bai, W.~Liu, {Revisiting Multiple Instance Neural Networks}, Pattern Recognition 74~(C) (2018) 15--24.

\bibitem{Ilse2020DeepMI}
M.~Ilse, J.~M. Tomczak, M.~Welling, {Deep Multiple Instance Learning for Digital Histopathology}, in: Handbook of Medical Image Computing and Computer Assisted Intervention, 2020, pp. 521--546.

\bibitem{Pinheiro2015}
P.~O. Pinheiro, R.~Collobert, {From Image-Level to Pixel-Level Labeling with Convolutional Networks}, in: Computer Vision and Pattern Recognition, 2015, pp. 1713--1721.

\bibitem{ilse2018attention}
M.~Ilse, J.~Tomczak, M.~Welling, {Attention-Based Deep Multiple Instance Learning}, in: International Conference on Machine Learning, 2018, pp. 2127--2136.

\bibitem{shao2021transmil}
Z.~Shao, H.~Bian, Y.~Chen, Y.~Wang, J.~Zhang, X.~Ji, et~al., {Transmil: Transformer Based Correlated Multiple Instance Learning for Whole Slide Image Classification}, in: Neural Information Processing Systems, 2021, pp. 2136--2147.

\bibitem{rymarczyk2021kernel}
D.~Rymarczyk, A.~Borowa, J.~Tabor, B.~Zielinski, {Kernel Self-Attention for Weakly-Supervised Image Classification Using Deep Multiple Instance Learning}, in: Winter Conference on Applications of Computer Vision, 2021, pp. 1721--1730.

\bibitem{zhou2012multi}
Z.~Zhi-Hua, Z.~Min-Ling, H.~Sheng-Jun, L.~Yu-Feng, {Multi-Instance Multi-Label Learning}, Artificial Intelligence 176~(1) (2012) 2291--2320.

\bibitem{yang2017miml}
H.~Yang, J.~Tianyi~Zhou, J.~Cai, Y.~Soon~Ong, {MIML-FCN+: Multi-Instance Multi-Label Learning via Fully Convolutional Networks with Privileged Information}, in: Computer Vision and Pattern Recognition, 2017, pp. 1577--1585.

\bibitem{feng2017deep}
J.~Feng, Z.-H. Zhou, {Deep MIML Network}, in: Association for the Advancement of Artificial Intelligence, 2017, pp. 1884--1890.

\bibitem{ren2023proposal}
H.~Ren, W.~Yang, T.~Zhang, Y.~Zhang, {Proposal-Based Multiple Instance Learning for Weakly-Supervised Temporal Action Localization}, in: Computer Vision and Pattern Recognition, 2023, pp. 2394--2404.

\bibitem{arnab2021vivit}
A.~Arnab, M.~Dehghani, G.~Heigold, C.~Sun, M.~Lu{\v{c}}i{\'c}, C.~Schmid, {Vivit: A Video Vision Transformer}, in: International Conference on Computer Vision, 2021, pp. 6836--6846.

\bibitem{lv2023unbiased}
H.~Lv, Z.~Yue, Q.~Sun, B.~Luo, Z.~Cui, H.~Zhang, {Unbiased Multiple Instance Learning for Weakly Supervised Video Anomaly Detection}, in: International Conference on Computer Vision, 2023, pp. 8022--8031.

\bibitem{chen2024prompt}
J.~Chen, L.~Li, L.~Su, Z.-j. Zha, Q.~Huang, {Prompt-Enhanced Multiple Instance Learning for Weakly Supervised Video Anomaly Detection}, in: Computer Vision and Pattern Recognition, 2024, pp. 18319--18329.

\bibitem{javed2022additive}
S.~A. Javed, D.~Juyal, H.~Padigela, A.~Taylor-Weiner, L.~Yu, A.~Prakash, {Additive Mil: Intrinsically Interpretable Multiple Instance Learning for Pathology}, Advances in Neural Information Processing Systems 35~(20689--20702).

\bibitem{qu2023boosting}
L.~Qu, Z.~Yang, M.~Duan, Y.~Ma, S.~Wang, M.~Wang, Z.~Song, {Boosting Whole Slide Image Classification from the Perspectives of Distribution, Correlation and Magnification}, in: International Conference on Computer Vision, 2023, pp. 21463--21473.

\bibitem{tang2023multiple}
W.~Tang, S.~Huang, X.~Zhang, F.~Zhou, Y.~Zhang, B.~Liu, {Multiple Instance Learning Framework with Masked Hard Instance Mining for Whole Slide Image Classification}, in: International Conference on Computer Vision, 2023, pp. 4078--4087.

\bibitem{chen2023rankmix}
Y.-C. Chen, C.-S. Lu, {Rankmix: Data Augmentation for Weakly Supervised Learning of Classifying Whole Slide Images with Diverse Sizes and Imbalanced Categories}, in: Proceedings of the IEEE/CVF Conference on Computer Vision and Pattern Recognition, 2023, pp. 23936--23945.

\bibitem{early2024inherently}
J.~Early, G.~Cheung, K.~Cutajar, H.~Xie, J.~Kandola, N.~Twomey, {Inherently Interpretable Time Series Classification via Multiple Instance Learning}, in: International Conference on Learning Representations, 2024.

\bibitem{Shiku_2025_WACV}
K.~Shiku, K.~Nishimura, D.~Suehiro, K.~Tanaka, R.~Bise, {Ordinal Multiple-Instance Learning for Ulcerative Colitis Severity Estimation with Selective Aggregated Transformer}, in: Winter Conference on Applications of Computer Vision, 2025, pp. 4290--4299.

\bibitem{schwab2022automatic}
E.~Schwab, G.~O. Cula, K.~Standish, S.~S. Yip, A.~Stojmirovic, L.~Ghanem, C.~Chehoud, {Automatic Estimation of Ulcerative Colitis Severity from Endoscopy Videos Using Ordinal Multi-Instance Learning}, Computer Methods in Biomechanics and Biomedical Engineering: Imaging \& Visualization 10~(4) (2022) 425--433.

\bibitem{ren2019pancreatic}
B.~Ren, X.~Liu, A.~A. Suriawinata, {Pancreatic Ductal Adenocarcinoma and Its Precursor Lesions: Histopathology, Cytopathology, and Molecular Pathology}, The American journal of pathology 189~(1) (2019) 9--21.

\bibitem{travis2011international}
W.~D. Travis, E.~Brambilla, M.~Noguchi, A.~G. Nicholson, K.~R. Geisinger, Y.~Yatabe, D.~G. Beer, C.~A. Powell, G.~J. Riely, P.~E. Van~Schil, et~al., {International Association for the Study of Lung Cancer/American Thoracic Society/European Respiratory Society International Multidisciplinary Classification of Lung Adenocarcinoma}, Journal of thoracic oncology 6~(2) (2011) 244--285.

\bibitem{mciver2017privacy}
A.~McIver, T.~Rabehaja, R.~Wen, C.~Morgan, {Privacy in Elections: How Small is “Small”?}, Journal of information security and applications 36 (2017) 112--126.

\bibitem{no_label_no_cry}
G.~Patrini, R.~Nock, P.~Rivera, T.~Caetano, {(Almost) No Label No Cry}, in: Neural Information Processing Systems, 2014, p. 190–198.

\bibitem{wang2011mixture}
Z.~Wang, L.~Lan, S.~Vucetic, {Mixture Model for Multiple Instance Regression and Applications in Remote Sensing}, IEEE Transactions on Geoscience and Remote Sensing 50~(6) (2011) 2226--2237.

\bibitem{shikuIcassp}
K.~Shiku, S.~Matsuo, D.~Suehiro, R.~Bise, {Counting Network for Learning from Majority Label}, in: International Conference on Acoustics, Speech and Signal Processing, 2024, pp. 7025--7029.

\bibitem{liu2024pseudo}
P.~Liu, L.~Ji, X.~Zhang, F.~Ye, {Pseudo-Bag Mixup Augmentation for Multiple Instance Learning-Based Whole Slide Image Classification}, IEEE Transactions on Medical Imaging 43~(5) (2024) 1841--1852.

\bibitem{gadermayr2023mixup}
M.~Gadermayr, L.~Koller, M.~Tschuchnig, L.~M. Stangassinger, C.~Kreutzer, S.~Couillard-Despres, G.~J. Oostingh, A.~Hittmair, {Mixup-Mil: Novel Data Augmentation for Multiple Instance Learning and a Study on Thyroid Cancer Diagnosis}, in: International Conference on Medical Image Computing and Computer-Assisted Intervention, 2023, pp. 477--486.

\bibitem{yang2022remix}
J.~Yang, H.~Chen, Y.~Zhao, F.~Yang, Y.~Zhang, L.~He, J.~Yao, {Remix: A General and Efficient Framework for Multiple Instance Learning Based Whole Slide Image Classification}, in: International Conference on Medical Image Computing and Computer-Assisted Intervention, 2022, pp. 35--45.

\bibitem{li2021novel}
Z.~Li, W.~Zhao, F.~Shi, L.~Qi, X.~Xie, Y.~Wei, Z.~Ding, Y.~Gao, S.~Wu, J.~Liu, Y.~Shi, D.~Shen, {A Novel Multiple Instance Learning Framework for COVID-19 Severity Assessment via Data Augmentation and Self-Supervised Learning}, Medical Image Analysis 69~(101978).

\bibitem{li2019deep}
M.~Li, L.~Wu, A.~Wiliem, K.~Zhao, T.~Zhang, B.~Lovell, {Deep Instance-Level Hard Negative Mining Model for Histopathology Images}, in: International Conference on Medical Image Computing and Computer-Assisted Intervention, 2019, pp. 514--522.

\bibitem{qu2022dgmil}
L.~Qu, X.~Luo, S.~Liu, M.~Wang, Z.~Song, {Dgmil: Distribution Guided Multiple Instance Learning for Whole Slide Image Classification}, in: International Conference on Medical Image Computing and Computer-Assisted Intervention, 2022, pp. 24--34.

\bibitem{as2024sm}
F.~M. Castro-Mac{\'\i}as, P.~Morales-Alvarez, Y.~Wu, R.~Molina, A.~Katsaggelos, {Sm: Enhanced Localization in Multiple Instance Learning for Medical Imaging Classification}, in: Annual Conference on Neural Information Processing Systems, 2024, pp. 77494--77524.

\bibitem{qu2022bi}
L.~Qu, M.~Wang, Z.~Song, et~al., {Bi-Directional Weakly Supervised Knowledge Distillation for Whole Slide Image Classification}, in: Advances in Neural Information Processing Systems, Vol.~35, 2022, pp. 15368--15381.

\bibitem{krizhevsky2009learning}
A.~Krizhevsky, G.~Hinton, {Learning Multiple Layers of Features from Tiny Images}.

\bibitem{netzer2011reading}
Y.~Netzer, T.~Wang, A.~Coates, A.~Bissacco, B.~Wu, A.~Y. Ng, {Reading Digits in Natural Images with Unsupervised Feature Learning}, NIPS workshop on deep learning and unsupervised feature learning.

\bibitem{yang2023medmnist}
J.~Yang, R.~Shi, D.~Wei, Z.~Liu, L.~Zhao, B.~Ke, H.~Pfister, B.~Ni, {MedMNIST V2-A Large-Scale Lightweight Benchmark for 2D and 3D Biomedical Image Classification}, Scientific Data 10~(1) (2023) 41.

\bibitem{Paszke2019PyTorchAI}
A.~Paszke, S.~Gross, F.~Massa, A.~Lerer, J.~Bradbury, G.~Chanan, T.~Killeen, Z.~Lin, N.~Gimelshein, L.~Antiga, A.~Desmaison, A.~K{\"o}pf, E.~Yang, Z.~DeVito, M.~Raison, A.~Tejani, S.~Chilamkurthy, B.~Steiner, L.~Fang, J.~Bai, S.~Chintala, {PyTorch: An Imperative Style, High-Performance Deep Learning Library}, in: Neural Information Processing Systems, 2019, pp. 8026 -- 8037.

\bibitem{he2016deep}
K.~He, X.~Zhang, S.~Ren, J.~Sun, {Deep Residual Learning for Image Recognition}, in: Computer Vision and Pattern Recognition, 2016, pp. 770--778.

\bibitem{adam}
K.~D. P, B.~Jimmy, {Adam: A Method for Stochastic Optimization}, arXiv preprint arXiv:1412.6980.

\bibitem{xu2024mctformer+}
L.~Xu, M.~Bennamoun, F.~Boussaid, H.~Laga, W.~Ouyang, D.~Xu, {Mctformer+: Multi-class Token Transformer for Weakly Supervised Semantic Segmentation}, IEEE Transactions on Pattern Analysis and Machine Intelligence 46 (2024) 8380--8395.

\bibitem{qu2022bidirectional}
L.~Qu, xiaoyuan Luo, M.~Wang, Z.~Song, {Bi-directional Weakly Supervised Knowledge Distillation for Whole Slide Image Classification}, in: Neural Information Processing Systems, 2022, pp. 15368--15381.

\end{thebibliography}

\clearpage

\begin{appendices}


\end{appendices}
\color{black}


\end{document}